\documentclass[11pt,a4paper]{article}
\usepackage[nohyperref]{emnlp2017}
\usepackage{times}
\usepackage{url}
\usepackage{latexsym}
\usepackage{expex}
\usepackage{booktabs}
\usepackage{todonotes}
\usepackage{amsmath}
\usepackage{enumitem}
\usepackage{calc}
\usepackage{caption}
\usepackage{subcaption}
\usepackage{multicol}

\lingset{interpartskip=0px,aboveexskip=1ex,belowexskip=1ex,labeloffset=1ex,textoffset=1ex}

\title{Room for improvement in automatic image description: an error analysis}

\author{Emiel van Miltenburg \\
  Vrije Universiteit Amsterdam \\
  {\tt emiel.van.miltenburg@vu.nl} \And
  Desmond Elliott \\
  ILLC, Universiteit van Amsterdam \\
  {\tt d.elliott@uva.nl}}

\emnlpfinalcopy
\date{}

\begin{document}
\maketitle

\begin{abstract}

In recent years we have seen rapid and significant progress in automatic image description but what are the open problems in this area? Most work has been evaluated using text-based similarity metrics, which only indicate that there have been improvements, without explaining what has improved. In this paper, we present a detailed error analysis of the descriptions generated by a state-of-the-art attention-based model. Our analysis operates on two levels: first we check the descriptions for accuracy, and then we categorize the types of errors we observe in the inaccurate descriptions. We find only 20\% of the descriptions are free from errors, and surprisingly that 26\% are unrelated to the image. Finally, we manually correct the most frequently occurring error types (e.g. gender identification) to estimate the performance reward for addressing these errors, observing gains of 0.2--1 BLEU point per type.
\end{abstract}
\section{Introduction}
Automatic image description is the task of describing an image in natural language \cite{bernardi2016automatic}. Recent advances in this area have been evaluated with
text-based similarity metrics, such as BLEU
\cite{Papineni:2002:BMA:1073083.1073135} or Meteor
\cite{denkowski:lavie:meteor-wmt:2014}. These metrics make it easy 
for researchers to benchmark the effect of their 
modeling decisions, but they are not informative about the strengths and weaknesses of a proposed model. This is especially true for n-gram based metrics, such as BLEU, which measure grammatical fluency and not semantic adequacy \cite{Reiter2009}.

In this paper, we present a coarse- and fine-grained analysis of the
descriptions generated by a state-of-the-art attention-based model
\cite{xu2015show}, trained on the Flickr30K dataset
\cite{young2014image}.
The goal of this paper is to assess the qualities of a state-of-the-art model to illustrate the recent progress in this area and the challenges ahead.
The coarse analysis quantifies whether the descriptions are
accurate or inaccurate, while the fine-grained analysis quantifies the
types of errors observed in the descriptions. We define accurate to mean that the description is congruent with the image, without it necessarily being the ``best'' or most complete description. We find that 80\% of the
descriptions contain at least one type of inaccuracy, and that 26\% are
completely wrong. In addition to categorizing the errors, we perform a manual error correction study to estimate the reward for addressing these errors. 
We find that fixing the five most frequently occurring errors contributes
between 0.2 -- 1 BLEU points improvement over the baseline, for each type of error. We hope that our findings will encourage future research to address the specific errors we have observed.\footnote{All our code, data, and annotation guidelines will be available upon publication.}

\section{Related work}

\begin{figure*}
\begin{subfigure}[t]{0.24\textwidth}
\centering
\includegraphics[width=1\linewidth,trim={0 10mm 0 10mm},clip]{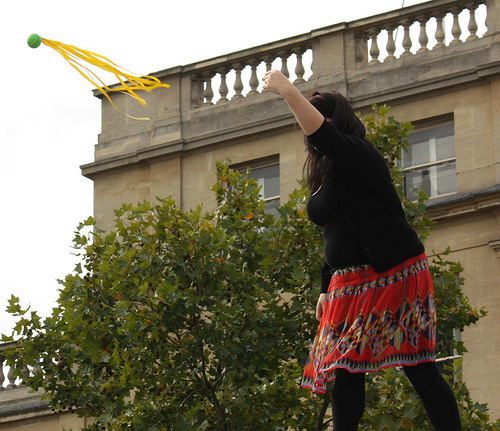}
\small A woman in a \textbf{red} shirt is standing in front of a building
\caption{One error}
\label{fig:one_error}
\end{subfigure}
\hfill%
\begin{subfigure}[t]{0.24\textwidth}
\centering
\includegraphics[width=1\linewidth]{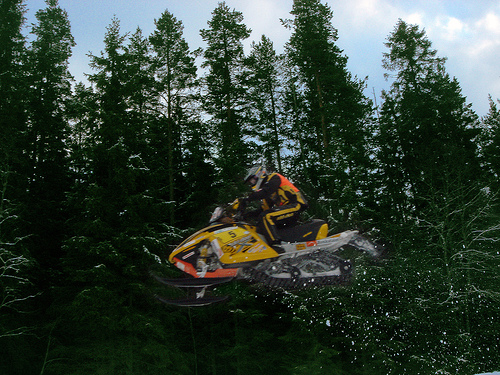}
\small A man in a \textbf{yellow} helmet rides a \textbf{bike} in the air
\caption{Two errors}
\end{subfigure}
\hfill%
\begin{subfigure}[t]{0.24\textwidth}
\centering
\includegraphics[width=1\linewidth]{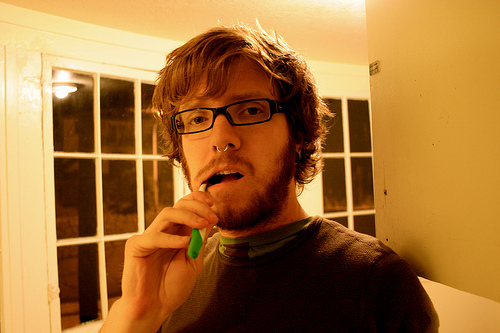}
\small A blond \textbf{woman} in a \textbf{white} shirt is \textbf{blowing} her teeth
\caption{Three errors}
\end{subfigure}
\hfill%
\begin{subfigure}[t]{0.24\textwidth}
\centering
\includegraphics[width=1\linewidth]{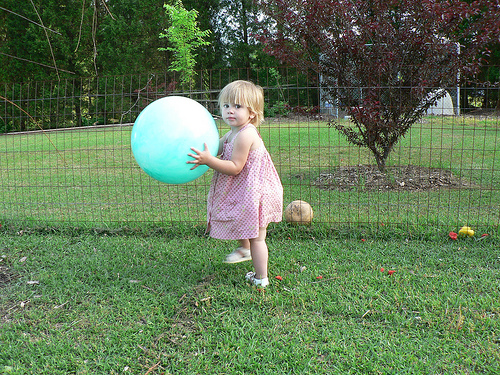}
\small A little \textbf{boy} in a \textbf{white shirt} playing \textbf{soccer} 
\caption{Four errors}
\label{fig:generally-unrelated}
\end{subfigure}
\caption{Examples of images with 1--4 errors. The annotated errors are marked in boldface.}
\end{figure*}

Early work on image description was evaluated with text-based similarity measures \emph{and} a human judgment study \cite{bernardi2016automatic}. This type of judgment study 
involves asking humans to rate whether the descriptions
accurately describe the image, are grammatically correct, are relevant for the image, are human-like, {\it inter-alia}, using a Likert-scale survey. The main criticisms of human judgment studies is they are expensive to perform and difficult to replicate without access to the same
subject pool and control samples (e.g. \citealt{Papineni:2002:BMA:1073083.1073135,hodosh2016focused}).
Nevertheless, these studies are the clearest indication of the differences between models. Our coarse-grained analysis is a binarized version of the correctness scale from \cite{mitchell2012midge}.

In this paper, our main focus is to provide a detailed analysis of the
quality of descriptions generated by a state-of-the-art model.
Our work is most closely related to \newcite{hodosh2016focused}, who propose an evaluation of image description systems using binary 
forced-choice tasks, where systems have to choose the best description
for a given image. For each image, the system can choose between the
original description or a manipulated description. By controlling the
manipulations, the authors are able to check for weaknesses in image
description systems.  Their error categories partially overlap with ours, though we
provide a more fine-grained typology.

\section{Error categories}\label{sec:errorcategories}

We developed a non-exhaustive categorisation of errors by inspecting the descriptions generated by an attention-based image description model \cite{xu2015show}. We trained the model on the Flickr30K dataset \cite{young2014image}, with 300D word embeddings, a 1000D GRU hidden layer \cite{Cho2014}, and `CONV$_{5,4}$' image features from the VGG-19 CNN \cite{Simonyan2015}. We generated 1,014 descriptions with a beam width of five hypotheses, recording a Meteor score of 17.4 on the Flickr30K test set.

In total, we identified 20 common types of errors, which we grouped into four main categories: \textsc{people, subject, object}, and \textsc{general}. 
We developed annotation guidelines with examples for each type of error. Due to space constraints, we provide the annotation guidelines in the supplementary material. The error categories and types of errors are described below.

\begin{description}[noitemsep, leftmargin=0cm, topsep=0px, itemindent=1.5\parindent]
\item[People] Image description models often make mistakes that are specific to the description of people. Types of errors in this category are \textsc{age} (e.g. \emph{woman} instead of \emph{girl}), \textsc{gender} (\emph{man} instead of \emph{woman}), \textsc{type of clothing} (\emph{shirt} instead of \emph{jacket}), and \textsc{color of clothing} (\emph{red shirt} instead of \emph{blue shirt}).
\item[Subject] Mistakes relating to the subject of the description. This category contains the following types of errors: \textsc{wrong} when the wrong entity in the image is chosen as the subject, \textsc{similar} when the model mis-identifies the subject for something visually similar (e.g. \emph{guitar} instead of \emph{violin}), \textsc{non-existent} when nothing close to the mentioned entity is present in the image, and \textsc{extra subject} when an additional (nonexistent) entity is described along with the correct entity.
\item[Object] Similar to {\bf Subject}.
\item[General] Mistakes that are not specific to people. Error types in this category are: \textsc{stance} for posture-related mistakes, \textsc{activity} for wrongly identified activities, \textsc{position} for mistakes in spatial relations within the image, \textsc{number} for counting errors (too few/many entities mentioned), \textsc{scene/event/location} for mis-identifications of the scene, event, or location, \textsc{color} for non-clothing entities that are mistakenly attributed with a color, \textsc{other} for any unforeseen mistakes, and \textsc{generally unrelated} for descriptions that do not seem to have any relation with the image. In these cases, it is impossible for annotators to assign any error category to the description. E.g.\ if Figure~\ref{fig:one_error} were to be described as \emph{A dog runs through the snow}.

\end{description}

\section{Annotation tasks}\label{sec:task-definitions}

We define two error annotation tasks: The \textbf{coarse-grained annotation} task is a binary categorization problem, where an annotator determines for every description whether it is accurate. The \textbf{fine-grained annotation} task is a multiclass categorization problem, given the error types presented in the previous section. Each inaccurate description is annotated with one or more error types. We can think of this task as a means to assess the \emph{semantic edit distance} between a generated description and the closest accurate alternative.

In total, one annotator categorized all 1,014 generated descriptions into the coarse-grained groups: accurate and inaccurate descriptions. The same annotator then performed the fine-grained annotation. We validated the annotation scheme by double-annotating a random selection of 100 descriptions (10\% of the data) to determine whether the annotation guidelines provide a reliable basis for annotating the errors.


\subsection{Results for the coarse-grained task}

In the coarse-grained annotation task, 812 out of 1014 descriptions (80\%) were judged to be inaccurate. We achieved a good inter-annotator agreement of Cohen's $\kappa$=0.67, with an accuracy of 91\%. The discrepancy between these numbers is explained by the label distribution: the \textsc{inaccurate} category is so dominant that any disagreement yields a high penalty in $\kappa$. Out of the 100 double-annotated descriptions, the first and second annotator judged 86 and 81 descriptions to be inaccurate, with agreement on 79 descriptions.

\subsection{Evaluating the fine-grained annotations}

In the fine-grained annotation task, we double-annotated the 79 descriptions that both annotators agreed contained at least one inaccuracy. Tables \ref{table:errorsperdescription}
 and \ref{table:errorcounts} show the number of errors per image, and the distribution of error types across the dataset. In total, we found 1,265 errors in 812 descriptions, which is an average of 1.56 errors / description. 
 
Surprisingly, the most common error category is \textsc{generally unrelated}  
(264 times). 
Errors from the \textsc{General} and \textsc{People} categories are much more frequent than the other two. Taken together, the \textsc{Subject} category is least common. Our intuition is that this is because mistakes in decoding the subject from the language model affect the entire sentence; the choice of subject influences the probability of all subsequent words, leading to a generally unrelated sentence.

The fine-grained annotation task is inherently ambiguous because inaccurate descriptions might be corrected in many different ways. Figure~\ref{fig:one_error} illustrates this ambiguity. The generated description for this image is given in Ex. (\ref{ex:ambiguous:original}). This description could either be corrected to (\ref{ex:ambiguous:alt1}) or (\ref{ex:ambiguous:alt2}), depending on whether one assumes the mistake is in the color or the type of clothing.

\pex %
\a \label{ex:ambiguous:original}A woman in a \textbf{red shirt} is standing in front of a building
\a \label{ex:ambiguous:alt1}A woman in a \textbf{black shirt} is standing \ldots
\a \label{ex:ambiguous:alt2}A woman in a \textbf{red skirt} is standing \ldots
\xe

Subjectivity and ambiguity are inherent to the task of image description; describing an image in one simple sentence means that you have to make a choice about what to include in your description. But this subjectivity also means that it is difficult to provide a proper intrinsic evaluation for the annotation task: different choices about how to describe an image may be equally valid. To quantify the extent of this issue, we treat the double annotation for the fine-grained task as a retrieval problem, i.e. how many error types are also found by the second annotator? This experiment achieves a precision of 0.54, with a recall of 0.55. Based on this observation, we decided to carry out an \emph{extrinsic} evaluation: how useful are the fine-grained annotations for guiding future research on model development? We discuss this evaluation below.

\begin{table}
\centering
\begin{tabular}{lrrrr}
\toprule
Errors 	   &   1 &   2 &  3 &  4 \\
Frequency & 486 & 221 & 83 & 22 \\
\bottomrule
\end{tabular}
\caption{The distribution of error annotations.}
\label{table:errorsperdescription}
\end{table}

\begin{table*}
\center
\begin{tabular}{lr}
\toprule
 Type                 &   Count \\
\midrule
 generally unrelated  &     264 \\
 color of clothing    &     195 \\
 activity             &     168 \\
 type of clothing     &     104 \\
 gender               &      98 \\
 scene/event/location &      91 \\
 number               &      61 \\
\bottomrule
\end{tabular}
\hspace{1ex}
\begin{tabular}{lr}
\toprule
 Type                 &   Count \\
\midrule
non-existent object   &      47 \\
 age                  &      40 \\
 stance               &      38 \\
 position             &      37 \\
 extra subject        &      34 \\
 similar-object       &      31 \\
 other                &      20 \\
 \bottomrule
 \end{tabular}
 \hspace{1ex}
\begin{tabular}{lrl}
\toprule
 Type                 &   Count \\
\midrule
 color                &      14 \\
 non-existent subject &      11 \\
 wrong-object         &       7 \\
 similar-subject      &       3 \\
 extra object         &       1 \\
 wrong-subject        &       1 \\
 \\
\bottomrule
\end{tabular}
\caption{Number of times each error was annotated in our fine-grained analysis.}
\label{table:errorcounts}
\end{table*}

\section{Correcting the errors}\label{sec:improving}

Now we have observed the frequency of each type of error, we can ask: would there be a positive effect  if a model could address these errors? We selected the five most common error types (excluding \textsc{generally unrelated}), and manually corrected each error \emph{without} looking at the reference descriptions. If a description is annotated with multiple errors, we only correct the relevant error. We tried to be conservative in our corrections; e.g. for \textsc{color of clothing} errors, if the system wrote e.g. \emph{white shirt} instead of \emph{checkered/leopard print/\ldots shirt}, we left the description untouched, rather than insert the pattern. For the \textsc{activity} errors, we tried to change as little as possible but editing the activity often also entails changing the object as well. For example, a sentence that read \emph{A man in a suit is holding a sign.} was changed to \emph{A man in a suit is talking.} because the man wasn't holding anything and leaving out the object would produce an ungrammatical sentence. If a change would entail completely re-ordering the sentence, we leave the generated description untouched.

Table~\ref{table:fixed} presents the BLEU and Meteor scores for the validation set before and after correction. For example, after only correcting the colors of clothing, we find a one-point improvement for the BLEU score with respect to the original model.

\begin{table}[h!]
\small
\centering
\begin{tabular}{lrrrr}
\toprule
 Type                 &   BLEU & $\Delta$   &   Meteor & $\Delta$   \\
\midrule
 Baseline             &   17.8 & ----    &     17.2 & ----    \\
 Color of clothing    &   18.8 & 1.0     &     17.5 & 0.3     \\
 Activity             &   18.5 & 0.7     &     17.7 & 0.5     \\
 Type of clothing     &   18.1 & 0.3     &     17.4 & 0.2     \\
 Gender               &   18.6 & 0.8     &     17.6 & 0.4     \\
 Scene/event/location &   18.0 & 0.2     &     17.4 & 0.2     \\
\bottomrule
\end{tabular}
\caption{Error categories and the BLEU-4 and Meteor scores after correcting the errors. $\Delta$ indicates improvement in the scores between the modified descriptions and the original descriptions.}
\label{table:fixed}
\end{table}

We did not investigate whether these effects are cumulative, i.e.\ what happens if we correct \emph{all} errors. Presumably, they are cumulative, but this task is not suitable for such an investigation because the corrections need to be restrictions in order for the improvement estimation to be accurate. If we allowed annotators to correct all the errors in a sentence, we would be giving them \emph{carte blanche} to rewrite everything, turning the analysis into an evaluation of human performance.

\section{Conclusion}
In this paper we provided an extensive error analysis for image descriptions generated by a state-of-the-art attention-based model.
Our main contributions are: 
(1) Providing a taxonomy of common errors in automatically generated image descriptions. 
(2) Quantifying the weaknesses of the model. We posit that any model with a similar architecture will have similar weaknesses. 
(3) Quantifying the possible improvement of this model if those weaknesses are addressed.

We focused on the nature of the inaccurate descriptions, and looked at different errors that these contain. But what about the accurate descriptions? The descriptions that \emph{are} accurate, are also much more general than the human descriptions, which usually include small, but salient details. We propose the following rule: if the majority of the human descriptions comments on an aspect of the image that is not addressed by a generated description, then that aspect could be improved. We plan to explore the consequences of this in future work.

We see two other perspectives to build on the observations from this paper. \textbf{Automated error analysis}: As noted earlier, \newcite{hodosh2016focused} carried out a study in which they evaluate image description models using binary forced-choice tasks, where models have to choose which description best describes a particular image. The choices are carefully manipulated, so that each task evaluates the model's performance in one area (e.g. recognizing scenes). Our taxonomy of errors could be used to extend the range of available tasks, for example with a task to evaluate the use of color terms; \textbf{Extending existing models}: Table~\ref{table:fixed} provides an indication of how much a model could improve by incorporating a dedicated module to detect color, actions, type of clothing, gender, and scenes. We expect that our work will encourage researchers in vision \& language to investigate this possibility. More generally, we hope that our taxonomy of error types will help others to go beyond similarity-based metrics, and to look at their model's output through a qualitative lens.
\section*{Acknowledgments}
EM is supported by the Netherlands Organization for Scientific Research (NWO) via the Spinoza-prize awarded to Piek Vossen (SPI 30-673, 2014-2019).  DE is supported by NWO Vici grant nr. 277-89-002 awarded to Khalil Sima'an.

\bibliography{references}
\bibliographystyle{emnlp_natbib}
\appendix
\section{Annotation Guidelines}

\subsection{Introduction}
This document provides guidelines for the annotation of automatically generated image descriptions. Our goal is to assess the semantic competence of image description models. In other words: are the descriptions at least `technically' correct? This is a low bar, as we ignore fluency and usefulness, which are also desirable properties for an NLG system. We define two tasks: 
\begin{enumerate}
\item \textbf{A binary decision task}, where annotators judge whether or not a description is congruent with an image.
\item \textbf{A categorization task}, where annotators select error categories that apply for incongruent descriptions.
\end{enumerate}

These tasks are strongly related: if a description is incongruent, it should fall into one of the error categories, and vice versa. Hence, annotators for either task need to be familiar with our taxonomy of errors.

\begin{table*}
\centering
\begin{tabular}{lllll}
\toprule
People & Subject & Object & General & General \\
\midrule
Age & Wrong & Wrong & Stance & Scene/event/location\\
Gender & Similar & Similar & Activity & Other\\
Type of clothing & Inexistent & Inexistent & Position & Color\\
Color of clothing & Extra subject & Extra object & Number & Generally unrelated\\
\bottomrule
\end{tabular}
\caption{Error categories for incongruent image descriptions. The organization of these categories corresponds to the organization of the categories in the annotation environment.}
\label{table:categories}
\end{table*}

\subsection{Error categories}
All our error categories are provided in Table~\ref{table:categories}. There are four main categories: People, Subject, Object, and General. I tried to strike a balance between specificity and amount of categories. No doubt some of these could be further subcategorized, but more categories means the annotation task might become overwhelming.

\subsubsection{Short description}
Here's a short description of each category, and each of the subcategories. The next subsection provides examples for each of these.
\begin{description}[noitemsep, leftmargin=0cm, topsep=0px, itemindent=1.5\parindent]
\item[People] Image description models often make mistakes that are specific to the description of people. Subcategories are \textsc{age} (e.g. \emph{woman} instead of \emph{girl}), \textsc{gender} (\emph{man} instead of \emph{woman}), \textsc{type of clothing} (\emph{shirt} instead of \emph{jacket}), and \textsc{color of clothing} (\emph{red shirt} instead of \emph{blue shirt}).
\item[Subject] Mistakes relating to the subject of the description. We use the following subcategories: \textsc{wrong} when the wrong entity in the image is chosen as the subject, \textsc{similar} when the image description system mis-identifies the subject for something visually similar (e.g. \emph{guitar} instead of \emph{violin}), \textsc{inexistent} when nothing close to the mentioned entity is present in the image, and \textsc{extra subject/object} when an additional (nonexistent) entity is mentioned besides the correct entity.
\item[Object] See \textbf{subject}.
\item[General] Mistakes that are not specific to people. The subcategories are as follows: \textsc{stance} for posture-related mistakes, \textsc{activity} for wrongly identified activities, \textsc{position} for mistakes in spatial relations within the image, \textsc{number} for any counting errors (too few/many entities mentioned), \textsc{scene/event/location} for misidentifications of the scene, event, or location, \textsc{color} for non-clothing entities that are mistakenly said to have a particular color, \textsc{other} for any unforeseen mistakes, and \textsc{generally unrelated} for generally unrelated descriptions, that are beyond repair. This is usually the case when more than 2--3 error (sub)categories are applicable.
\end{description}

\subsubsection{Examples}

\begin{center}

\includegraphics[trim={0 75mm 0 0},clip,width=130px]{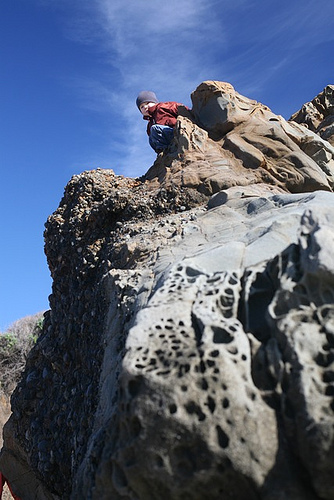}\\
A \textbf{man} is climbing a rock\\
Category: Age\\[3ex]

\includegraphics[width=130px]{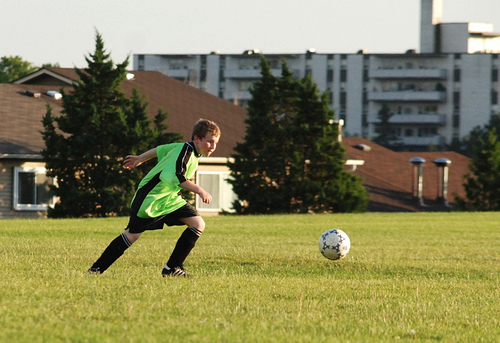}\\
A \textbf{girl} playing soccer\\
Category: Gender\\[3ex]

\includegraphics[width=130px]{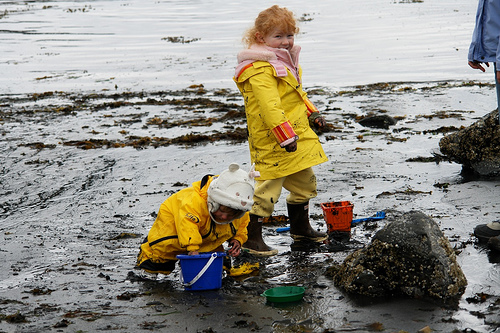}\\
A girl in a yellow \textbf{shirt} is standing on the beach\\
Category: Type of clothing\\[3ex]

\includegraphics[trim={0 20mm 0 10mm},clip,width=130px]{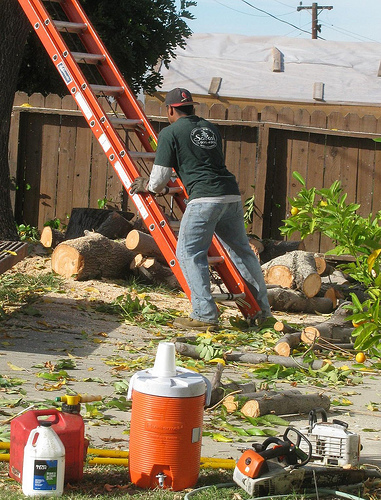}\\
A man in a \textbf{blue} shirt and blue jeans is working on a ladder\\ 
Category: Color of clothing\\[3ex]

\includegraphics[trim={0 35mm 0 20mm},clip,width=130px]{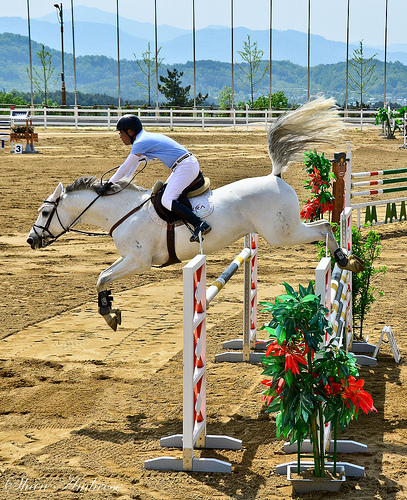}\\
A \textbf{boy} jumps over a hurdle\\
Category: Wrong subject\\[3ex]

\includegraphics[trim={0 20mm 0 10mm},clip,width=130px]{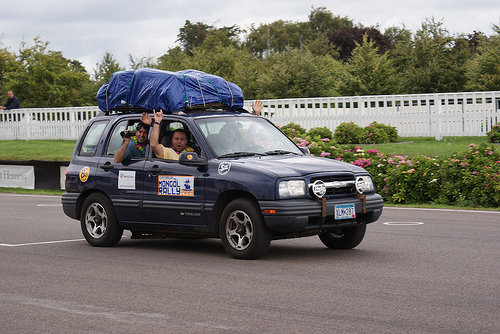}\\
\textbf{A woman in a blue shirt} is standing in front of a blue car\\
Category: Inexistent subject\\[3ex]

\includegraphics[trim={0 50mm 0 0mm},clip,width=130px]{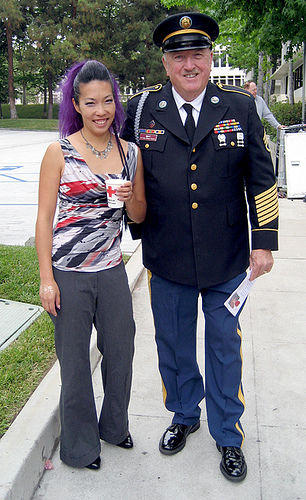}\\
\textbf{Two police officers} are posing for a picture\\
Category: Similar subject, number\\[3ex]

\includegraphics[trim={0 50mm 0 35mm},clip,width=130px]{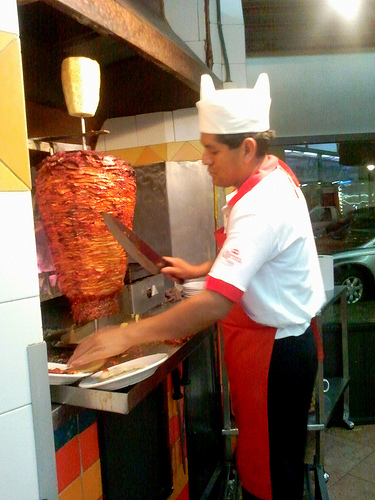}\\
A man in a white shirt \textbf{and a man in a white shirt} are preparing food\\ 
Category: Extra subject\\[3ex]

\includegraphics[trim={0 25mm 0 0mm},clip,width=130px]{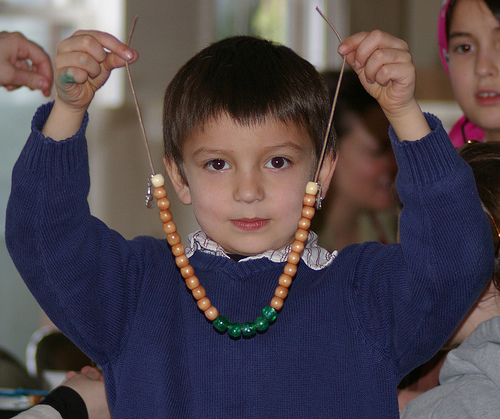}\\
A young boy is holding a \textbf{little girl}\\
Category: Wrong object\\[3ex]

\includegraphics[trim={0 90mm 0 20mm},clip,width=130px]{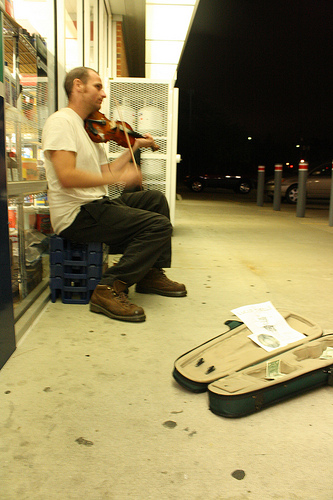}\\
A man is playing a \textbf{guitar}\\
Category: Similar object\\[3ex]

\includegraphics[width=130px]{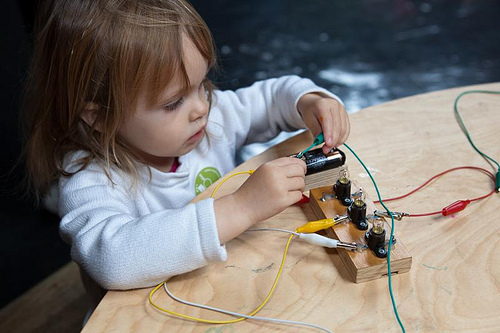}\\
A young girl in a white shirt is playing with a \textbf{guitar}\\ 
Category: Inexistent object\\[3ex]

\includegraphics[width=130px]{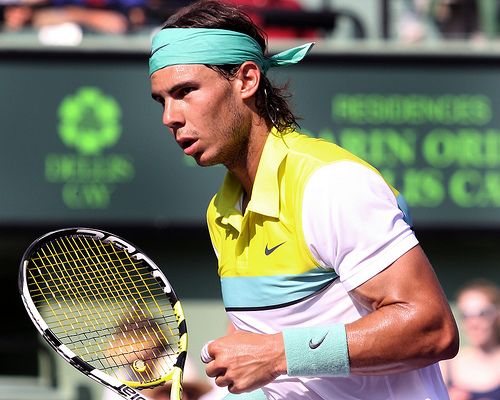}\\
A man with a tennis racket \textbf{and a tennis racket}\\ 
Category: Extra object\\[3ex]

\includegraphics[width=130px]{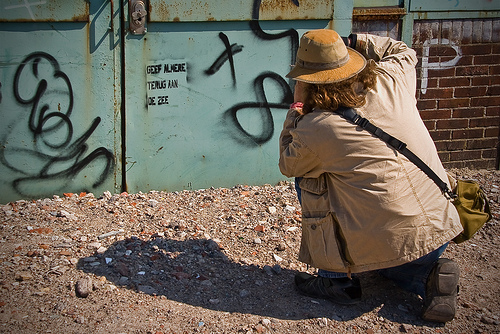}\\
A man in a brown jacket is \textbf{standing} in front of a wall\\ 
Category: Stance\\[3ex]

\includegraphics[width=130px]{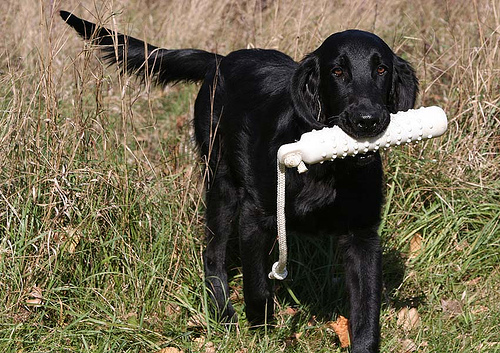}\\
A black dog \textbf{runs through} the grass\\ 
Category: Activity\\[3ex]

\includegraphics[width=130px]{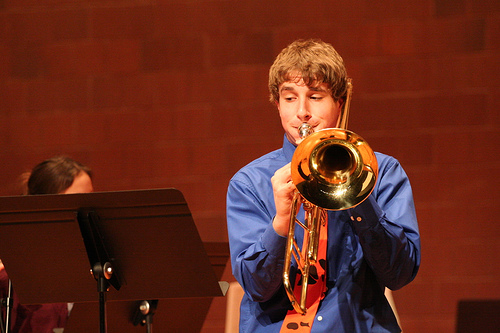}\\
\textbf{Two} men are playing instruments\\
Category: Number\\[3ex]

\includegraphics[trim={0 50mm 0 0mm},clip,width=130px]{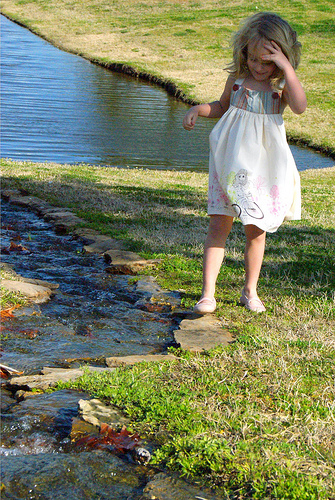}\\
A little girl in a white dress is walking \textbf{in} the water\\ 
Category: Position\\[3ex]

\includegraphics[width=130px]{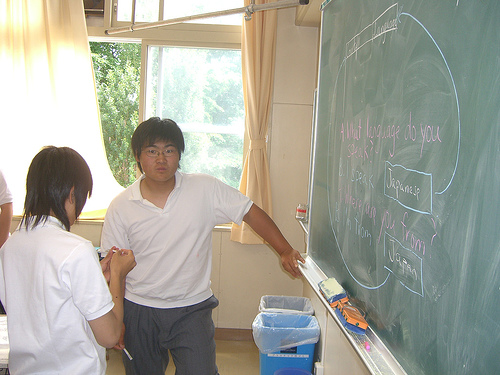}\\
A man in a white shirt and a woman in a white shirt are standing \textbf{in a hallway}\\ 
Category: Scene/event/location\\[3ex]

\includegraphics[width=130px]{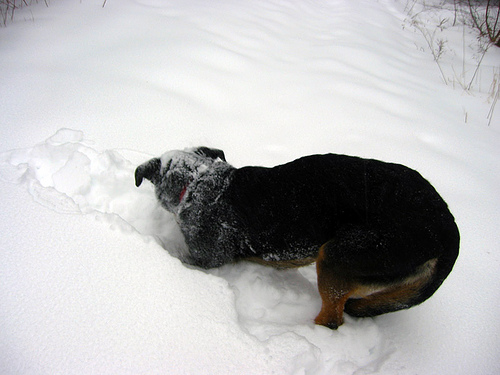}\\
A black \textbf{and white} dog is playing in the snow\\
Category: Color\\[3ex]

\includegraphics[width=130px]{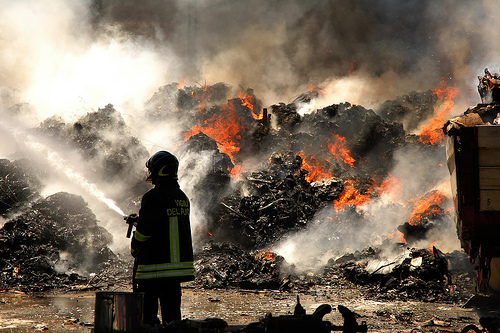}\\
\textbf{A group of people standing in the snow}\\ 
Category: Generally unrelated\\[3ex]

\includegraphics[width=130px]{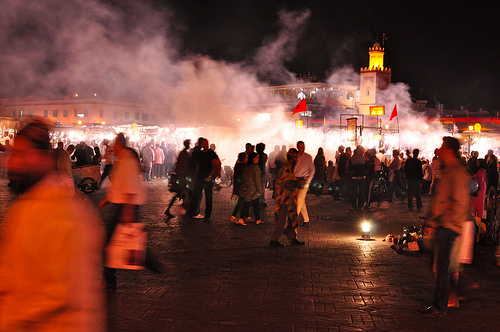}\\
A group of people are standing in \textbf{a fire}\\
Category: Other
\end{center}

\subsubsection{Important contrasts}
While the categories are fairly straightforward, there are cases where it is easy to get confused between a pair of categories. Here are additional guidelines for difficult cases that I have encountered.

\begin{itemize}
\item \textsc{stance} versus \textsc{activity}: Use the former when the difference is static, e.g. \emph{standing} vs. \emph{sitting}. Use the latter if the difference is dynamic, e.g. \emph{standing} versus \emph{walking}.

\item \textsc{scene/event/location} versus \textsc{position}: Use the former when the surroundings are not correct. Use the latter when position within the surroundings is not correct.

\item \textsc{extra subject/object} versus \textsc{number}: Use the former when the subject/object is wrongfully extended with a conjunction (e.g. \emph{and a woman in a white shirt}). Use the latter when there's a general mismatch in number (\emph{a, one, two, three, a group of}).

\item \textsc{similar object} versus \textsc{position}: This conflict arises in cases where e.g. \emph{\ldots is sitting on a bench} is used instead of \emph{\ldots is sitting on a chair}. In all these cases, use \emph{similar object}. (Even if there is an actual bench in the image.)
\end{itemize}

\subsection{Task descriptions \& instructions}
Now that we have seen the different error categories, we can describe the two main tasks as follows:

\begin{description}
\item[Task 1: Congruency] Judge whether the generated description is congruent (no error categories apply) or incongruent (at least one error category applies).

\item[Task 2: Categorizing incongruent descriptions] Annotate the `semantic edit distance' between the generated description and the closest valid description that you can imagine. Tick all the error categories corresponding to the things you would have to change. If the generated description is unrelated to the image, or if you feel that there are too many changes necessary to get to a valid description, select \textsc{generally unrelated}.
\end{description}

The threshold for when a description is generally unrelated is undefined. In general, I feel like type/color of clothing don't really hurt the relation between description and image as much as e.g. having the wrong verb. So it all comes down to your intuition.

\subsection{Evaluation: correcting the errors}
This is a separate task that serves both as an evaluation of Task 2, and as an indication of system performance if all errors identified in Task 2 are addressed. The correction task works as follows.

\begin{enumerate}
\item Select an error type to correct. E.g. \textsc{Color of clothing}. 
\item Go through all images annotated with this type, and correct \emph{only} the relevant error.
\item When all relevant errors are corrected, we evaluate the results using BLEU/Meteor.
\end{enumerate}

It is important for this task to be conservative in editing the descriptions. Try to change as little as possible. If a change would require restructuring the entire sentence, leave the description as it is. We'd rather underestimate than overestimate the improvement from fixing the errors. Otherwise we'd just be evaluating how good humans are at writing descriptions. So e.g.\ for colors, \emph{only} change color terms into other color terms. For gender, only change \emph{man $\leftrightarrow$ woman} and \emph{boy $\leftrightarrow$ girl}, not \emph{man $\leftrightarrow$ girl}. That would be changing the age along with the gender.

\end{document}